\documentclass[letterpaper, 10 pt, conference]{ieeeconf}
\usepackage{graphicx}
\usepackage{amsmath,amssymb}
\usepackage{mathtools}
\usepackage{xeCJK}
\usepackage{cite}
\usepackage{picinpar}
\usepackage{flushend}
\usepackage{colortbl}
\usepackage{soul}
\usepackage{multirow}
\usepackage{pifont}
\usepackage{amsfonts}  
\usepackage{color}
\usepackage{xcolor}
\usepackage{alltt}
\usepackage{enumerate}
\usepackage{siunitx}
\usepackage{pbox}
\usepackage{url}
\usepackage{mathrsfs}
\usepackage[hidelinks,breaklinks=true]{hyperref}
\usepackage[ruled,vlined,linesnumbered]{algorithm2e}
\DontPrintSemicolon
\SetKwInput{KwIn}{Inputs}
\SetKwInput{KwOut}{Outputs}
\SetKwComment{tcp}{\(\triangleright\) }{}
\SetAlFnt{\small}
\SetAlCapFnt{\small}
\SetAlCapNameFnt{\small}
\SetAlgoNlRelativeSize{-1}
\SetKw{Sample}{Sample}
\SetKw{Compute}{Compute}
\SetKw{Update}{Update}
\SetKw{Encode}{Encode}
\SetKw{Decode}{Decode}
\usepackage{booktabs} 

\IEEEoverridecommandlockouts                             
\title{\LARGE \bf
M$^{2}$GRPO: Mamba-based Multi-Agent Group Relative Policy Optimization  for Biomimetic Underwater Robots Pursuit
}

\author{Yukai Feng*, Zhiheng Wu*, Zhengxing Wu, \emph{Senior Member, IEEE}, \\ Junwen Gu, Junzhi Yu, \emph{Fellow, IEEE},  Min Tan
\thanks{*These authors contributed equally.}
\thanks{This work was supported in part by the National Natural Science Foundation of China under Grant 62433021, 62373353, and in part by Youth Innovation Promotion Association CAS (2023039).
\emph{(Corresponding author: Zhengxing~Wu.)}}
\thanks{Y. Feng, Z. Wu, J. Gu and M. Tan are with the School of Artificial Intelligence, University of Chinese Academy of Sciences, Beijing 100049, China, and also with the Key Laboratory of Cognition and Decision Intelligence for Complex Systems, Institute of Automation, Chinese Academy of Sciences, Beijing 100190, China  (e-mail:~{\tt\small fengyukai2021@ia.ac.cn; zhengxing.wu@ia.ac.cn; gujunwen2022@ia.ac.cn; min.tan@ia.ac.cn}). }%
\thanks{Z. Wu is with Baidu Inc., Beijing 100085, China (e-mail:~{\tt\small wzh404.ai@gmail.com}).}%
\thanks{J. Yu is with the State Key Laboratory for Turbulence and Complex Systems, Department of Advanced Manufacturing and Robotics, College of Engineering, Peking University, Beijing 100871, China  (e-mail:~{\tt\small junzhi.yu@ia.ac.cn}). }%
}

\begin{document}

\maketitle
\thispagestyle{empty}
\pagestyle{empty}

\begin{abstract}
Traditional policy learning methods in cooperative pursuit face fundamental challenges in biomimetic underwater robots, where long-horizon decision making, partial observability, and inter-robot coordination require both expressiveness and stability. 
To address these issues, a novel framework called Mamba-based multi-agent group relative policy optimization (M$^{2}$GRPO) is proposed, which integrates a selective state-space Mamba policy with group-relative policy optimization under the centralized-training and decentralized-execution (CTDE) paradigm.
Specifically, the Mamba-based policy leverages observation history to capture long-horizon temporal dependencies and exploits attention-based relational features to encode inter-agent interactions, producing bounded continuous actions through normalized Gaussian sampling.
To further improve credit assignment without sacrificing stability, the group-relative advantages are obtained by normalizing rewards across agents within each episode and optimized through a multi-agent extension of GRPO, significantly reducing the demand for training resources while enabling stable and scalable policy updates.
Extensive simulations and real-world pool experiments across team scales and evader strategies demonstrate that M$^{2}$GRPO consistently outperforms MAPPO and recurrent baselines in both pursuit success rate and capture efficiency. 
Overall, the proposed framework provides a practical and scalable solution for cooperative underwater pursuit with biomimetic robot systems.
\end{abstract}

\section{Introduction}
In recent research, biomimetic underwater robots, inspired by the propulsion and sensing mechanisms of marine organisms such as cetaceans and fish, have attracted increasing attention due to their high maneuverability, low noise, and effective stealth~\cite{cui2023review, wang2023versatile}. Building on these advantages, they have demonstrated wide applicability in domains such as resource exploration~\cite{li2023bioinspired}, equipment inspection~\cite{xie2024study}, search-and-rescue operations~\cite{iguchi2024agile}, and ecological monitoring~\cite{berlinger2021implicit}.
Compared to single-robot deployments, swarms of biomimetic robots further enhance efficiency and operational safety by leveraging wider coverage and collaborative decision-making, rendering them particularly suitable for complex aquatic environments featuring dynamic conditions and disturbances~\cite{yang2021survey, cai2023cooperative}.
Among the many cooperative tasks, pursuit–evasion (PE) stands out as a representative benchmark for demonstrating multi-robot interaction and collaboration, as it naturally incorporates challenges such as long-horizon decision-making, partial observability and relational coupling~\cite{antonio2024approximate}.
Consequently, it has emerged as an important touchstone for evaluating and advancing the cooperative capabilities of underwater multi-agent systems. Moreover, the related technologies hold great utility in both military and civilian domains, including missile interception, aircraft control, search and rescue, and beyond~\cite{xi2023zero, liu2023novel, lozano2022visibility}.
\par

\begin{figure*}[t]
  \centering
  \includegraphics[width=1.0\textwidth]{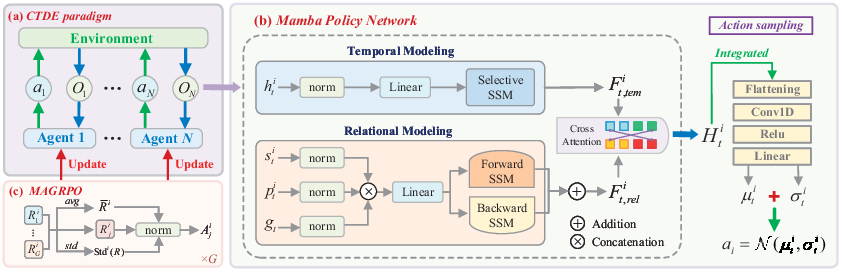}
  \caption{Overall framework of the proposed M$^{2}$GRPO algorithm, which consists of three components: 
  (a) CTDE paradigm: centralized training with decentralized execution, where agents share environment information and update in parallel during training, but rely solely on local observations and history for independent decision-making at the execution stage; 
  (b) Mamba Policy: a selective state-space architecture that models temporal dependencies with Mamba2 and relational features with BiMamba2, then fuses them via multi-head attention to produce expressive latent features; 
  (c) MAGRPO: a multi-agent extension of group-relative policy optimization, where group-normalized advantages across parallel environments are combined with a clipped PPO objective for stable and efficient policy updates.}
  \label{fig1}
\end{figure*}
Existing research on multi-robot PE task can be broadly divided into two categories. 
The first comprises geometric or control priors, such as dynamic game theory~\cite{xu2022multiplayer}, particle swarm optimization~\cite{geng2021particle}, and graph-theoretic modeling~\cite{zeng2020comparison}.
Although effective in specific scenarios, these methods rely heavily on task-specific assumptions and predefined model specifications, which substantially limit their robustness and scalability. 
In contrast, the second category is composed of policy-learning approaches based on multi-agent reinforcement learning (MARL), which facilitate both cooperation and competition within the centralized training with decentralized execution (CTDE) paradigm through shared or individualized policies~\cite{zhang2022multi, kouzeghar2023multi, de2021decentralized}. Representative algorithms include multi-agent deep deterministic policy gradient~\cite{wan2021improved, yang2023large} and multi-agent proximal policy optimization~\cite{lin2025distributed}.
While deep reinforcement learning avoids the need for precise system modeling and demonstrates strong adaptability across diverse environments, most existing studies continue to rely on multilayer perceptrons (MLPs) as the backbone for policy and value networks, which exhibits significant limitations in practical applications~\cite{xie2023recurrent}.
Specifically, such networks struggle to integrate historical information in partially observable and long-horizon tasks, often resulting in short-sighted decision-making.
Moreover, in multi-agent systems with dynamically evolving interaction topologies, their ability to capture inter-agent relationships remains severely constrained. 
Consequently, there is an urgent need for backbone architectures that can simultaneously model long temporal dependencies and naturally incorporate multi-agent interaction dynamics.\par

In recent years, Mamba has garnered considerable attention for its selective state space model (SSM) paradigm and linear-time sequential recursion~\cite{gu2023mamba, zhang2025marl}. 
Through the combination of selective scanning and input-dependent gating mechanisms, it not only produces robust representations of long-range dependencies but also maintains low computational overhead and high deployment efficiency. 
These properties make it particularly suitable for resource constrained scenarios or tasks that require long-sequence reasoning. Unlike memoryless static MLP mappings, recurrent selective state updates in Mamba enable the effective integration of historical cues, thereby capturing long-term dependencies in partially observable and long-horizon tasks~\cite{dao2024transformers}. Furthermore, compared with Transformers that rely on quadratic-complexity self-attention, its linear-time recurrence substantially reduces inference latency and resource consumption, which better aligns with the demands of online recursion and onboard deployment~\cite{ozccelik2024chemical}. Consequently, Mamba offers a structural solution to long-horizon credit assignment under partial observability, while mitigating training instability from non-stationary interactions and enabling unified policies that capture both temporal and interactive dynamics.\par

However, the introduction of temporal backbones with enhanced expressivity tends to introduce computational burden, which poses new challenges to the efficiency of reinforcement learning frameworks.
Drawing inspiration from recent advances in large-scale models, DeepSeek-R1, which leverages group relative policy optimization (GRPO) to reduce computational cost and improve stability without explicit value networks~\cite{shao2024deepseekmath, guo2025deepseek, mroueh2025reinforcement}, this work introduces multi-agent group relative policy optimization (MAGRPO) for efficient policy training.
The algorithm extends GRPO from natural language generation without environment interaction to multi-agent reinforcement learning with simulation interaction.
It optimizes policies using episode-level returns, generates parallel trajectories from identical initial states, and derives relative advantages from group-averaged returns, which are then updated through PPO-clip.
This design enables MAGRPO to preserve training stability without the requirement for value networks, ultimately reducing the computational complexity and implementation cost of multi-agent training.\par
In summary, this paper proposes a Mamba-based multi-agent group relative policy optimization (M$^{2}$GRPO) algorithm as illustrated in Fig.~\ref{fig1}, which enhances the policy learning capability of underwater robot fish in collaborative pursuit tasks. The main contributions of this paper are as follows:
\begin{itemize}
    \item A Mamba-based policy network is proposed which leverages selective state-space modeling to capture long-term temporal dependencies from local observation histories, while incorporating attention-enhanced interaction features to represent dynamic relational features within a unified interaction framework.
    \item An extension of GRPO to multi-agent scenarios is developed which employs group-normalized advantage estimation to eliminate the need for explicit value functions and complex baselines, thereby improving training stability and scalability while significantly reducing computational resource requirements.
    \item Extension simulation and real-world experiments have verified the effectiveness of the proposed framework, demonstrating the first integration of a Mamba-based decentralized policy with multi-agent group-relative optimization for underwater multi-robot decision-making and providing new perspectives on intelligent cooperation in underwater environments.
\end{itemize}
\begin{figure}[!t]
  \centering
  \includegraphics[scale=0.9]{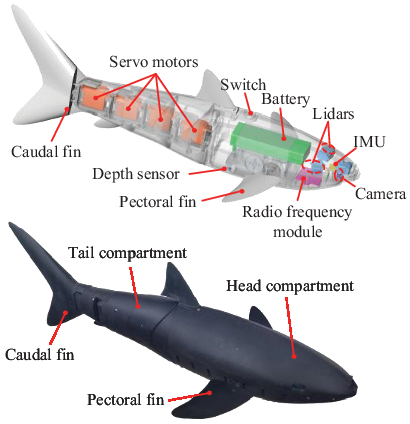}
  \caption{Mechatronic design and physical prototype of the biomimetic robot shark}
  \label{fig2}
\end{figure}
\vspace{-6pt}

\section{Preliminaries}
\subsection{Robotic Fish Platform}
Fig.~\ref{fig2} illustrates the prototype of the biomimetic robot shark, whose external morphology is inspired by \textit{Negaprion brevirostris}. The robot measures 0.68~m in length and has a mass of 3.3~kg, with detailed structural parameters reported in our previous work~\cite{yan2022real}.
Similar to the real shark, the robotic shark employs a multi-joint caudal fin to perform periodic oscillations, thereby generating both thrust and yaw torque. This enables excellent underwater maneuverability for executing tasks such as target tracking and evasive maneuvers. 
The control of caudal fin oscillation is driven by a central pattern generator (CPG) model that outputs rhythmic signals to a PWM controller for servo actuation, ensuring smooth and coordinated tail-beating. The swimming behavior is primarily modulated by the flapping frequency $\omega$ and offset $b$. Additionally, prior work~\cite{feng2025decentralized} achieved policy transfer by designing state-transition equations tailored to the motion characteristics of the biomimetic robot shark and demonstrated its effectiveness. The same approach is adopted in this work, with methodological details omitted and available in the cited reference.\par
\subsection{Problem Formulation}
\begin{figure}[!t]
  \centering
  \includegraphics[scale=0.9]{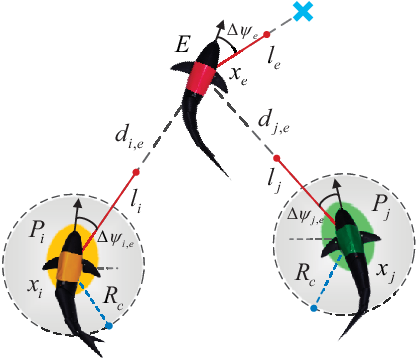}
  \caption{Illustration of the pursuit–evasion task with two pursuers $P_i$, $P_j$ and one evader $E$. Each pursuer is assigned a perception range $R_c$. The distance between the pursuit and evader is denoted as $d_{i,e}$.}
  \label{fig3}
\end{figure}

This study addresses a multi-biomimetic robot shark PE task in a two-dimensional bounded environment, where less maneuverable pursuers must coordinate efficiently and adapt their strategies adaptively to capture the agile evader.
As illustrated in Fig.~\ref{fig3}, the pursuer team comprises two robotic sharks with a capture radius of $R_c$ and a maximum swimming speed of $0.2$~m/s, while the evader possesses greater maneuverability than the pursuers, which maximum swimming speed is $0.3$~m/s.
In each episode, the pursuit is considered successful if the distance $d_{i,j}$ between any pursuer $P_i$ and the evader $E$ is less than $R_c$.
The core of this task is to design a distributed decision-making network that enables the pursuers to leverage cooperative advantages, compensate for individual mobility deficiencies, and achieve efficient capture.\par
\section{Method}
\subsection{Mamba Policy}
In this study, each agent adopts a Mamba Policy as its policy network. This architecture is built upon selective state-space modeling, enabling the agent to effectively capture long-term temporal dependencies from observation histories while extracting task-relevant features for decision making. At time step $t$, the local observation of agent $i$ is defined as
\begin{align}
\small
o_t^i = [s_t^i, g_t, p_t^i], \notag
\end{align}
where $s_t^i$ denotes the self-state information, $g_t$ represents the evader information, and $p_t^i$ corresponds to the collective features, which include the relative positions and velocities of neighboring pursuers with respect to agent $i$. The policy network takes two inputs: (i) the current local observation $o_t^i \in \mathbb{R}^{d}$, where $d$ is the observation dimension, and (ii) the historical observation sequence $h_t^i = [o_{t-L}^i, \cdots ,o_t^i] \in \mathbb{R}^{L \times d}$, where $L$ is the history length. Based on these inputs, it outputs the parameters of a Gaussian distribution, consisting of the mean $\mu_t^i$ and standard deviation $\sigma_t^i$, from which the continuous action $a_t^i$ is obtained through sampling and clipping.\par

To capture temporal dependencies, the Mamba Policy first applies a linear transformation to the historical observations $h_t^i$, which are fed into the Mamba2 module. Built on selective state-space modeling, this module leverages input-dependent gating and segmented recursion to efficiently capture long-range temporal features with linear time complexity, producing the temporal feature:
\begin{align}
F^i_{t,\text{tem}} = \mathrm{Mamba2}\!\left(\mathrm{Linear}\!\left(h_t^i\right)\right)
\end{align}
During batch training, temporal sequences of varying length are padded to a uniform size, with a padding mask applied to prevent invalid segments from affecting feature extraction.\par

In addition, the current observation $o_t^i$ is decomposed into the self-state vector $s_t^i$, the target feature $g_t$, and the collective features $p_t^i$. After linear mapping and normalization, these components are concatenated and passed to the BiMamba2 module, which leverages bidirectional temporal recursion to extract interaction features:
\begin{align}
F^i_{t,\text{rel}} = \mathrm{BiMamba2}\!\big(\mathrm{Concat}(s_t^i,\, g_t,\, p_t^i)\big)
\end{align}
resulting in the relational feature vector $F^i_{t,\text{rel}}$.
Subsequently, a multi-head attention mechanism is applied to integrate temporal and relational features:
\begin{align}
\small
H_t^i=\mathrm{MHA}\Big(F^i_{t,\text{rel}},\,F^i_{t,\text{tem}},\, F^i_{t,\text{tem}};\,M^i_{t,\text{tem}}\Big)\,,
\end{align}
where the relational features serve as the query, temporal features act as the key-value pairs, and $M^i_{t,\text{tem}}$ denotes the padding mask for historical observations. The fused representation $H_t^i$ is then passed through fully connected layers to generate the parameters of the action distribution:
\begin{align}
&\mu_{t}^{i}=\operatorname{tanh}(W_{\mu}H_t^i)\cdot\alpha \notag \\ 
&\sigma_{t}^{i}=\mathrm{softplus}(W_{\sigma}H_t^i)+\eta,
\end{align}
where $W_\mu$ and $W_\sigma$ are linear projection matrices, $\alpha$ is an amplitude scaling factor, and $\eta = 1 \times 10^{-3}$ is a degeneration suppression term. Finally, the continuous action $a_t^i$ is obtained from a Gaussian distribution ${a_i} = {\cal N}(\mu _t^i,\sigma _t^i)$ through sampling.\par

\subsection{Policy Improvement via MAGRPO}
In the training process, a multi-agent group relative policy optimization (MAGRPO) algorithm is proposed, which extends GRPO to the multi-agent setting. Specifically, the system involves $N$ agents with each agent $i$ performing the same task in $G$ parallel environments. Let $R_j^i$ denote the average episodic return of agent $i$ in the $j$-th environment. The group-normalized advantage for agent $i$ is defined as:
\begin{align}
A_j^i = \frac{R_j^i - \overline{R}^i}{\mathrm{Std}^i(R) + \tau}, \quad j=1,\cdots,G,
\end{align}
where $\overline{R}^i = \frac{1}{G}\sum_{j=1}^G R_j^i$ is the average return of agent $i$ across $G$ environments, $\mathrm{Std}^i(R)$ denotes the standard deviation of returns, and $\tau$ is a stabilization coefficient.
By normalizing returns across agents within each parallel group, this design standardizes the advantage estimates, thereby eliminating the need for explicit value functions or global baselines, and ensuring stable credit assignment under multi-environment sampling.\par

Subsequently, a PPO-based clipped objective is employed to update the policy. Let $\pi_\theta(a_t^i|o_t^i,h_t^i)$ denote the current policy and $\pi_{\theta_{\text{old}}}$ is the previous policy. The optimization objective is defined as:
{\small
\begin{align}
L^{i}(\theta)
&= \mathbb{E}_{j,t}\!\left[\min\!\Big(\rho_{j,t}^{i}(\theta)\,A_{j}^{i},\;
\mathrm{clip}\!\big(\rho_{j,t}^{i}(\theta),\,1-\epsilon,\,1+\epsilon\big)\,A_{j}^{i}\Big)\right], \notag \\
\rho_{j,t}^{i}(\theta)
&= \frac{\pi_{\theta}\!\big(a_{j,t}^{i}\mid o_{j,t}^{i},h_{j,t}^{i}\big)}
{\pi_{\theta_{\text{old}}}\!\big(a_{j,t}^{i}\mid o_{j,t}^{i},h_{j,t}^{i}\big)},
\end{align}}
where $\rho_{j,t}^i(\theta)$ denotes the probability ratio of agent $i$ at time $t$ in environment $j$, and $\epsilon$ is the clipping threshold. By maximizing this objective, the policy can be improved while constraining updates within a trust region for stable learning.
During centralized training, agents share environment information and update policies in parallel to promote coordination. During execution, each agent acts independently based only on its local observations and history, enabling decentralized execution. Although no centralized critic is used, this CTDE paradigm preserves scalability and supports real-world deployment.\par

\subsection{Reward Design}
In this study, the instantaneous reward $r^i$ obtained by agent $i$ during interaction with the environment is composed of three terms:
\begin{equation}
\begin{aligned}
  r^i =  r_{cap}^{i} + r_{aux}^{i} + r_{safe}^{i},
\end{aligned}
\end{equation}
where $r_{cap}^{i}$ is the capture reward, $r_{aux}^{i}$ is the auxiliary guidance reward, and $r_{safe}^{i}$ is the safety reward.
Meanwhile, reward coefficients and thresholds are selected by scale balancing with small-range pilot sweeps, and team coordination is still enforced via the team-terminating capture event under CTDE centralized value estimation although $r^i$ is defined per agent. \par

Specifically, the capture reward $r_{cap}^{i}$ provides a fixed value of 12 whenever the Euclidean distance $d_{i,j}$ between a pursuer and the evader falls below the capture threshold $R_c$. The auxiliary reward $r_{aux}^{i}$ encourages the pursuer to converge toward the evader and is defined as:
\begin{equation}
\begin{aligned}
  r_{aux}^{i} = -0.35 \| \boldsymbol{x}_i - \boldsymbol{x}_{e} \|,
\end{aligned}
\end{equation}
where $\boldsymbol{x}_i$ and $\boldsymbol{x}_e$ denote the position vectors of the pursuer and the evader, respectively. In addition, to ensure that the task proceeds within a safe range, a safety reward $r_{safe}^{i}$ is introduced to penalize boundary-violating behaviors:
\begin{equation}
\begin{aligned}
  r_{safe}^{i} =
  \begin{cases}
  0,  &  |d_i| < 1.85~\text{m} \\
  80 (|d_i| - 1.85),  &  1.85~\text{m} \leq |d_i| < 2~\text{m} \\
  \min(\text{e}^{2|d_i| - 4},  12),  &  |d_i| \geq 2~\text{m}
  \end{cases}
\end{aligned}
\end{equation}
where $|d_i|$ represents the distance between agent $i$ and the map boundary. This penalty increases progressively as the agent approaches the boundary, thereby constraining its motion range and preventing unsafe behaviors.\par

\subsection{Algorithm and Implementation Details}
Combining the proposed policy network, optimization method, and reward design, the training procedure and key parameter settings are presented.
Accordingly, the training is conducted with $G=10$ parallel environments over $E=600$ episodes, each with a horizon of $T=25$.
Policy optimization is based on PPO-clip, with the clipping threshold of $\epsilon=0.2$. 
The Adam optimizer is employed with a learning rate of $1\times 10^{-3}$ and gradient norm clipping to enhance training stability.
The policy network uses a hidden dimension of 64. In the action head, an amplitude scaling factor $\alpha = 2$ is applied to regulate the output range, and a degeneration suppression parameter $\eta = 10^{-3}$ is introduced 
to ensures numerical stability. Gaussian noise is added for exploration and linearly annealed from 0.5 to 0, facilitating a smooth shift from exploration to exploitation. The complete training procedure is summarized in Algorithm~\ref{alg:m2grpo}, and the evader strategy is pre-trained using the DDPG method.

\begin{algorithm}[t]
\caption{M$^{2}$GRPO}
\label{alg:m2grpo}
\SetAlgoLined\DontPrintSemicolon
\SetKwInOut{KwIn}{Input}
\KwIn{Environment $\mathcal{E}$, policies $\{\pi_{\theta_i}\}_{i=1}^N$, 
episodes $E$, horizon $T$, parallel envs $G$, agents $N$, 
update iters $K$, stab $\gamma$}

\For{$\mathrm{ep}=1 \cdots E$}{
  Reset environments;\;
  Initialize histories $h_0^i$ and returns $R_g^i$\;
    \For{$t=1 \cdots T$}{
      Each agent $i$ samples action $a_t^i \sim \pi_{\theta_i}(\cdot \mid o_t^i,h_t^i)$\;
      $(o_{t+1},\{r_t^i\}_{i=1}^N) \leftarrow \mathcal{E}.\mathrm{step}(\{a_t^i\}_{i=1}^N)$\;
      Update returns $R_g^i \leftarrow R_g^i + r_t^i$;\;
      Store $(o_t^i,h_t^i,a_t^i)$
    }
    
  \For{$g=1 \cdots G$ and $i=1 \cdots N$}{
    Compute mean return $\bar R_g^i \!\leftarrow\! R_g^i/T$\;
    Compute group-relative advantage 
    $A_g^i \!\leftarrow\! \dfrac{\bar R_g^i - \overline R_g}{\mathrm{Std}_g(R)+\tau}$\;
  }
  
  \For{$i=1 \cdots N$}{
    \textit{Freeze old policy snapshot}: $\pi_{\theta_{\text{old}},i} \leftarrow \pi_{\theta_i}$\;
    \For{$k=1 \cdots K$}{
      Compute ratio $\rho_{g,t}^i = \dfrac{\pi_{\theta_i}(a_{g,t}^i \mid o_{g,t}^i,h_{g,t}^i)}{\pi_{\theta_{\text{old}},i}(a_{g,t}^i \mid o_{g,t}^i,h_{g,t}^i)}$\;
      Update objective with PPO-clip:
     $\mathcal{L} \leftarrow \mathcal{L} - \text{PPO-clip loss}( \rho_{g,t}^i, A_g^i )$;\;
      Update parameters $\theta_i$\;
    }
  }
}
\end{algorithm}

\section{Experiments and Discussion}
In this section, the proposed M$^{2}$GRPO algorithm is validated through both simulations and real-world experiments. In simulations, M$^{2}$GRPO is evaluated against three representative baselines, and ablation studies examine the contribution of key modules. In real-world experiments, the learned policy is deployed on biomimetic robot shark platforms to verify its feasibility and effectiveness. During testing, the evader adopts either a random strategy or a learned strategy as detailed in~\cite{feng2025cooperative}.

\subsection{Simulations and Quantitative Evaluation}
In the comparative experiments, three mainstream baselines are selected to comprehensively evaluate the effectiveness of the proposed M$^{2}$GRPO algorithm. Their brief descriptions and hyperparameter settings are summarized as follows:

\begin{figure}[!t]
  \centering
  \includegraphics[scale=0.9]{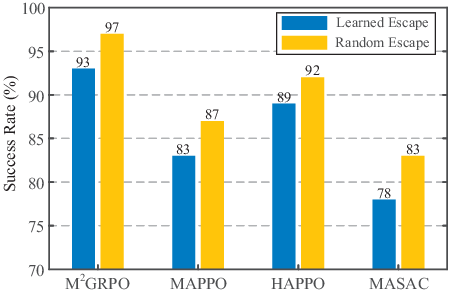}
  \caption{Capture success rate of pursuers under different evader strategies: (i) evader with a learned policy; (ii) evader with a random policy.}
  \label{fig4}
\end{figure}

\begin{figure}[!t]
  \centering
  \includegraphics[scale=0.9]{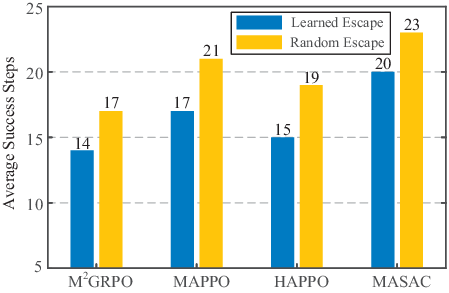}
  \caption{Average steps to successful capture under different evader strategies: (i) evader with a learned policy, (ii) evader with a random policy.}
  \label{fig5}
\end{figure}

\begin{figure}[!t]
  \centering
  \includegraphics[scale=0.9]{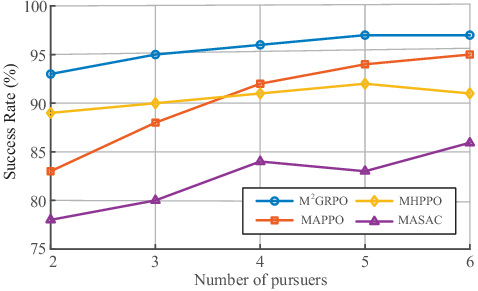}
  \caption{The success rate of the pursuit for different numbers of pursuers.}
  \label{fig6}
\end{figure}

\begin{itemize}
\item \textbf{MAPPO (Multi-agent PPO)}: A policy gradient algorithm with centralized critics and decentralized execution, and serves as a strong baseline in multi-agent scenarios. Key hyperparameters include: clipping threshold $\epsilon=0.2$, GAE decay factor $\lambda=0.95$, discount factor $\gamma=0.99$, Adam optimizer with a learning rate of $1\times 10^{-3}$, and $K=20$ policy update epochs.
\item \textbf{HAPPO (Heterogeneous-agent PPO)}: A policy optimization method with per-agent trust-region constraints, designed to improve the stability of policy updates and widely regarded as a strong baseline. Key hyperparameters include: trust-region step limit $\delta=0.01$, discount factor $\gamma=0.99$, GAE decay factor $\lambda=0.95$, Adam optimizer with a learning rate of $1\times 10^{-3}$, and $K=20$ policy update epochs.
\item \textbf{MASAC (Multi-agent soft actor-critic)}: An entropy-regularized multi-agent algorithm that enhances exploration and demonstrates robustness in both sparse-reward and high-noise environments. Key hyperparameters include: discount factor $\gamma=0.99$, temperature coefficient $\alpha=0.2$, Adam optimizer with a learning rate of $1\times 10^{-3}$, and batch size $256$.
\end{itemize}
To validate policy performance, evaluation is conducted in terms of effectiveness, efficiency, and scalability. In particular, effectiveness is measured by the success rate of cooperative pursuit, efficiency by the average number of steps to capture, and scalability by varying the number of pursuers.
To ensure statistical reliability, the reported results are averaged over 100 independent trials, with agent initial positions randomized at the beginning of each trial.\par

\subsubsection{Effectiveness}
Fig.~\ref{fig4} illustrates the comparison between four methods under two evader strategies in terms of success rate. Among the compared methods, M$^{2}$GRPO attains overall success rates of 97\% and 93\% in the two scenarios and outperforms all baselines. 
This advantage is primarily attributed to its joint modeling of temporal and relational features, which enables more effective coordination among pursuers and enhanced forecasting of the evader’s motion intent. 
Among the baselines, HAPPO emerges as the closest competitor to M$^{2}$GRPO by enforcing per-agent trust-region constraints, which ensure monotonic policy improvement and stable training. However, the absence of long-horizon temporal modeling leads to a slight reduction in overall success rate.
In addition, MAPPO leverages centralized value estimation to promote cooperation and achieves effective pursuit. Nevertheless, the lack of memory units makes it susceptible to unstable training and suboptimal convergence.
As an off-policy entropy-regularized approach, MASAC enables sufficient exploration during the early stage of training. However, its reliance on target entropy introduces residual randomness in later stages. Combined with the non-stationarity of multi-agent environments and noisy value estimates, its success rate is comparatively lower.\par

\subsubsection{Efficiency}
In the experiments shown in Fig.~\ref{fig5}, the average capture steps across 100 trials are reported. Benefiting from group relative policy optimization, M$^{2}$GRPO demonstrates more forward-looking planning of pursuit strategies, thereby requiring fewer steps on average in all scenarios and reflecting higher decision-making efficiency.
For the baselines, HAPPO enforces per-agent trust-region constraints that ensure stable capture. However, the absence of explicit long-term memory or sequence modeling restricts its predictive capability, leading to slightly higher step counts. In comparison, MAPPO relies on centralized value estimation to foster cooperative strategies. However, without temporal and relational encoders and under diluted credit assignment, it often produces suboptimal moves and redundant decisions, which further increase the step count.
With MASAC, residual randomness and noise in off-policy value estimation reduce its decisiveness and cause it to require the largest number of steps to achieve capture.

\subsubsection{Scalability}
Fig.~\ref{fig6} illustrates the impact of scaling pursuers from two to six on success rate.
The results demonstrate that by integrating multi-head attention over temporal and relational features to efficiently extract long-horizon history and inter-agent interactions, M$^{2}$GRPO consistently achieves the highest success rate by maintaining strong representational capacity and coordinated decision-making as the number of agents increases.
In comparison, MAPPO demonstrates strong scalability by leveraging parameter sharing, while the incorporation of a stable clipping mechanism and generalized advantage estimation within the policy optimization framework further enhances its robustness, leading to performance that ranks second only to M$^{2}$GRPO.
By contrast, the non-sharing variant MHPPO incurs linearly increasing training overhead and model size as the number of agents grows, resulting in weaker scalability than the former two methods.
Finally, MASAC assigns each agent an independent actor coupled with a centralized critic, which exacerbates the challenges of credit assignment and function approximation as the agent population scales, thereby inducing training instability and sample inefficiency, and ultimately resulting in the weakest overall performance.
\begin{table}[t]
\centering
\caption{Ablation results for the 2-pursuer vs. 1-evader task.}
\label{tab:ablation}
\small
\resizebox{\columnwidth}{!}{%
\begin{tabular}{lcccc}
\toprule
\textbf{Variant} & \textbf{Temporal} & \textbf{Relational} & \textbf{Backbone} & \textbf{Rate (\%)}~$\boldsymbol{\Delta}$ \\
\midrule
Full (M$^{2}$GRPO + Mamba) & \checkmark & \checkmark & Mamba & 92~~(0) \\
M$^{2}$GRPO + Mamba \textit{no history} & $\times$ & \checkmark & Mamba & 86~~($-6$) \\
M$^{2}$GRPO + Mamba \textit{no relation} & \checkmark & $\times$ & Mamba & 81~~($-11$) \\
M$^{2}$GRPO + MLP & $\times$ & $\times$ & MLP & 66~~($-26$) \\
\bottomrule
\end{tabular}}
\end{table}

\begin{figure*}[t]
  \centering
  \includegraphics[scale=0.9]{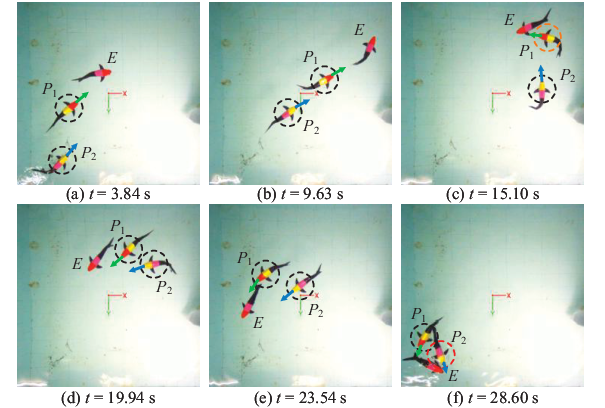}
  \caption{Snapshots of the cooperative pursuit experiment for bionic underwater robots}
  \label{fig7}
\end{figure*}
\begin{figure}[!t]
  \centering
  \includegraphics[scale=0.9]{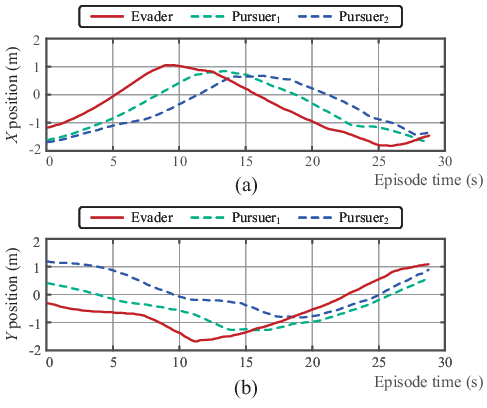}
  \caption{Positional relationship between the evader and the pursuers.}
  \label{fig8}
\end{figure}
\subsection{Ablation study}
To assess the contribution of each component, we conduct the ablation study by progressively removing or modifying modules: (i) dropping the temporal modeling branch, (ii) removing interaction encoding, and (iii) replacing the Mamba backbone with a plain MLP. 
As shown in Table~\ref{tab:ablation}, the complete model (M$^{2}$GRPO+Mamba) achieves the strongest performance. 
Notably, among the variants, removing interaction encoding leads to the most significant degradation, demonstrating the essential role of relational encoders. 
Furthermore, disabling temporal modeling also leads to performance degradation, confirming the advantage of leveraging observation history for long-horizon prediction.
In addition, substituting the selective state-space (Mamba) backbone with a standard MLP leads to further decline, highlighting the importance of the backbone’s temporal expressivity.
In summary, the ablation study confirms that temporal modeling and interaction encoding are critical for cooperation, and that group-relative optimization with the Mamba backbone achieves promising performance.

\subsection{Experiments and Qualitative Analysis}
To evaluate the effectiveness of the learned policy, underwater PE experiments are conducted using three biomimetic robot sharks. The experiments are carried out in an indoor pool with dimensions of $4~\text{m} \times 4~\text{m}$. Each robotic shark is equipped with a radio frequency communication module and a motion control unit to enable real-time information reception and processing. At the beginning of each trial, the initial positions and orientations of all 
robots are randomly assigned, with the coordinate origin defined at the pool center.
Fig.~\ref{fig7} illustrates the entire process of a cooperative pursuit during the experiment. The movement directions of the pursuers are indicated by arrows, with the green one denoting pursuer $P_1$ and the blue one denoting pursuer $P_2$. The capture radius $R_c$ of each pursuer is depicted as a black circle. When the distance between a pursuer and the evader satisfies $d_{i,e} \leq 0.3~\text{m}$, the circle turns red, indicating a successful capture. In contrast, when the distance lies within $0.3~\text{m} < d_{i,e} \leq 0.5~\text{m}$, the circle turns yellow, signaling that the capture is imminent.\par

For clarity, one representative experiment is presented to analyze the general characteristics of the learned policy. 
In the initial stage, as illustrated in Figs.~\ref{fig7}(a)--(b), pursuers $P_1$ and $P_2$ employed a straightforward pursuit strategy to rapidly approach the evader. As the task progressed, the evader is gradually driven toward the corner of the pool. 
At the stage illustrated in Fig.~\ref{fig7}(c), the sensing circle of pursuer $P_1$ turns yellow, which indicates that a capture is imminent.
However, the evader leveraged its high maneuverability to escape from the corner, as illustrated in Fig.~\ref{fig7}(d).
In the subsequent stage, Fig.~\ref{fig7}(e) demonstrates that pursuers $P_1$ and $P_2$ adjusted their strategy, transitioning from direct chasing to a two-sided encirclement to initiate the second pursuit. Ultimately, as shown in Fig.~\ref{fig7}(f), the evader is once again confined to the corner region and is successfully captured by pursuer $P_2$.\par
\begin{figure}[!t]
  \centering
  \includegraphics[scale=0.9]{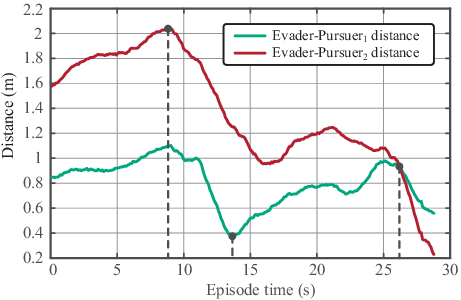}
  \caption{Distance variation between the evader and the pursuers.}
  \label{fig9}
\end{figure}
For quantitative analysis, Fig.~\ref{fig8} and Fig.~\ref{fig9} illustrate the variations of the agents’ positions along the $x$-axis and $y$-axis, as well as the distance relationships between pursuers and the evader. As shown in Fig.~\ref{fig8}, the pursuer team initially attempted to approach the evader along both $x$-axis and $y$-axis. 
However, shown in Fig.~\ref{fig9}, due to the evader’s superior maneuverability, the direct pursuit strategy is unable to reduce the distance before $9.3~\text{s}$.
As the evader is gradually cornered, the distance between pursuer $P_1$ and the evader decreased and reached a minimum at $14.5~\text{s}$, which indicates a high likelihood of capture.
Nevertheless, the evader successfully escaped by exploiting its maneuverability and adaptive evasion strategy, extending the distance to the pursuers. At $26~\text{s}$, the pursuer team adjusted its strategy, with $P_1$ shifting to an interception role while $P_2$ carried out the capture attempt. 
In the final stage, through cooperative coordination, the trajectory of the evader is encircled along the $x$-axis and continuous compression along the $y$-axis by the pursuers, which resulted in a successful capture. 
This process demonstrates the feasibility and adaptability of the learned pursuit strategy in dynamic environments.\par

\section{Conclusion and Future Work}\label{sec:6}
In this work, a novel framework, M$^{2}$GRPO, is proposed, which integrates a Mamba-based temporal-relational policy network with a group-relative optimization scheme under the CTDE paradigm, enabling efficient and stable training with both expressive representational capacity and high training efficiency.
Specifically, the Mamba policy captures long-horizon temporal dependencies and encodes relational features, while multi-head attention fuses these representations and maps them to Gaussian distributions for generating continuous and bounded actions.
On the optimization side, M$^{2}$GRPO extends group relative policy optimization to multi-agent settings by normalizing trajectory rewards across parallel groups, enabling efficient credit assignment without explicit value functions while preserving PPO-clip stability.
The effectiveness of the proposed framework has been validated through extensive simulations and real-world experiments, where M$^{2}$GRPO consistently outperforms strong baselines. 
Furthermore, ablation studies confirm the necessity of both temporal and relational modeling, with the Mamba backbone playing a pivotal role in performance gains.
Overall, this work proposes a novel training framework for cooperative operations of biomimetic underwater robots, which holds significant application value for underwater inspection, environmental exploration, and emergency response.\par
Future research will extend the algorithm to fully three-dimensional aquatic environments, integrate onboard perception and communication for end-to-end policy learning, and explore model-based or world-model reinforcement learning to enhance sample efficiency.


\begin{thebibliography}{99}

\bibitem{cui2023review}
Z. Cui, L. Li, Y. Wang, Z. Zhong, and J. Li, ``Review of research and control technology of underwater bionic robots,'' \emph{Intell. Mar. Technol. Syst.}, vol. 1, no. 1, p. 7, 2023.

\bibitem{wang2023versatile}
T. Wang, H.-J. Joo, S. Song, W. Hu, C. Keplinger, and M. Sitti, ``A versatile jellyfish-like robotic platform for effective underwater propulsion and manipulation,'' \emph{Sci. Adv.}, vol. 9, no. 15, p. eadg0292, 2023.

\bibitem{li2023bioinspired}
G. Li, T.-W. Wong, B. Shih, C. Guo, L. Wang, J. Liu, T. Wang, X. Liu, J. Yan, B. Wu, \emph{et al.}, ``Bioinspired soft robots for deep-sea exploration,'' \emph{Nat. Commun.}, vol. 14, no. 1, p. 7097, 2023.

\bibitem{xie2024study}
O. Xie, C. Zhang, C. Shen, Y. Li, and D. Zhou, ``Study on the hydrodynamic performance of a self-propelled robot fish swimming in pipelines environment,'' \emph{Ocean Eng.}, vol. 309, p. 118356, 2024.

\bibitem{iguchi2024agile}
K. Iguchi, T. Shimooka, S. Uchikai, Y. Konno, H. Tanaka, Y. Ikemoto, and J. Shintake, ``Agile robotic fish based on direct drive of continuum body,'' \emph{npj Robot.}, vol. 2, no. 1, p. 7, 2024.

\bibitem{berlinger2021implicit}
F. Berlinger, M. Gauci, and R. Nagpal, ``Implicit coordination for 3D underwater collective behaviors in a fish-inspired robot swarm,'' \emph{Sci. Robot.}, vol. 6, no. 50, p. eabd8668, 2021.

\bibitem{yang2021survey}
Y. Yang, Y. Xiao, and T. Li, ``A survey of autonomous underwater vehicle formation: Performance, formation control, and communication capability,'' \emph{IEEE Commun. Surv. Tutor.}, vol. 23, no. 2, pp. 815--841, 2021.

\bibitem{cai2023cooperative}
W. Cai, Z. Liu, M. Zhang, and C. Wang, ``Cooperative artificial intelligence for underwater robotic swarm,'' \emph{Robot. Auton. Syst.}, vol. 164, p. 104410, 2023.

\bibitem{antonio2024approximate}
E. Antonio, I. Becerra, and R. Murrieta-Cid, ``Approximate methods for visibility-based pursuit-evasion,'' \emph{IEEE Trans. Robot.}, early access, 2024.

\bibitem{xi2023zero}
A. Xi, Y. Cai, Y. Deng, and H. Jiang, ``Zero-sum differential game guidance law for missile interception engagement via neuro-dynamic programming,'' \emph{Proc. Inst. Mech. Eng., Part G: J. Aerosp. Eng.}, vol. 237, no. 14, pp. 3352--3366, 2023.

\bibitem{liu2023novel}
W. Liu, J. Hu, H. Zhang, M. Y. Wang, and Z. Xiong, ``A novel graph-based motion planner of multi-mobile robot systems with formation and obstacle constraints,'' \emph{IEEE Trans. Robot.}, vol. 40, pp. 714--728, 2023.

\bibitem{lozano2022visibility}
E. Lozano, I. Becerra, U. Ruiz, L. Bravo, and R. Murrieta-Cid, ``A visibility-based pursuit-evasion game between two nonholonomic robots in environments with obstacles,'' \emph{Auton. Robots}, vol. 46, no. 2, pp. 349--371, 2022.

\bibitem{xu2022multiplayer}
Y. Xu, H. Yang, B. Jiang, and M. M. Polycarpou, ``Multiplayer pursuit-evasion differential games with malicious pursuers,'' \emph{IEEE Trans. Autom. Control}, vol. 67, no. 9, pp. 4939--4946, 2022.

\bibitem{geng2021particle}
N. Geng, Z. Chen, Q. A. Nguyen, and D. Gong, ``Particle swarm optimization algorithm for the optimization of rescue task allocation with uncertain time constraints,'' \emph{Complex Intell. Syst.}, vol. 7, no. 2, pp. 873--890, 2021.

\bibitem{zeng2020comparison}
X. Zeng, L. Yang, Y. Zhu, and F. Yang, ``Comparison of two optimal guidance methods for the long-distance orbital pursuit-evasion game,'' \emph{IEEE Trans. Aerosp. Electron. Syst.}, vol. 57, no. 1, pp. 521--539, 2020.

\bibitem{zhang2022multi}
Z. Zhang, X. Wang, Q. Zhang, and T. Hu, ``Multi-robot cooperative pursuit via potential field-enhanced reinforcement learning,'' in \emph{Proc. IEEE Int. Conf. Robot. Autom. (ICRA)}, 2022, pp. 8808--8814.

\bibitem{kouzeghar2023multi}
M. Kouzeghar, Y. Song, M. Meghjani, and R. Bouffanais, ``Multi-target pursuit by a decentralized heterogeneous UAV swarm using deep multi-agent reinforcement learning,''  in \emph{Proc. IEEE Int. Conf. Robot. Autom. (ICRA)}, 2023, pp. 3289--3295.

\bibitem{de2021decentralized}
C. De Souza, R. Newbury, A. Cosgun, P. Castillo, B. Vidolov, and D. Kulić, ``Decentralized multi-agent pursuit using deep reinforcement learning,'' \emph{IEEE Robot. Autom. Lett.}, vol. 6, no. 3, pp. 4552--4559, 2021.

\bibitem{wan2021improved}
K. Wan, D. Wu, Y. Zhai, B. Li, X. Gao, and Z. Hu, ``An improved approach towards multi-agent pursuit--evasion game decision-making using deep reinforcement learning,'' \emph{Entropy}, vol. 23, no. 11, p. 1433, 2021.

\bibitem{yang2023large}
H. Yang, P. Ge, J. Cao, Y. Yang, and Y. Liu, ``Large scale pursuit-evasion under collision avoidance using deep reinforcement learning,'' in \emph{Proc. IEEE/RSJ Int. Conf. Intell. Robots Syst. (IROS)}, 2023, pp. 2232--2239.

\bibitem{lin2025distributed}
Y. Lin, H. Gao, and Y. Xia, ``Distributed pursuit--evasion game decision-making based on multi-agent deep reinforcement learning,'' \emph{Electronics}, vol. 14, no. 11, p. 2141, 2025.

\bibitem{xie2023recurrent}
S. Xie, Z. Zhang, H. Yu, and X. Luo, ``Recurrent prediction model for partially observable MDPs,'' \emph{Inf. Sci.}, vol. 620, pp. 125--141, 2023.

\bibitem{gu2023mamba}
A. Gu and T. Dao, ``Mamba: Linear-time sequence modeling with selective state spaces,'' \emph{arXiv preprint arXiv:2312.00752}, 2023.

\bibitem{zhang2025marl}
R. Zhang, Y. Sun, Z. Zhang, J. Li, X. Liu, A. H. Fan, H. Guo, and P. Yan, ``MARL-MambaContour: Unleashing multi-agent deep reinforcement learning for active contour optimization in medical image segmentation,'' \emph{arXiv preprint arXiv:2506.18679}, 2025.

\bibitem{dao2024transformers}
T. Dao and A. Gu, ``Transformers are SSMs: Generalized models and efficient algorithms through structured state space duality,'' \emph{arXiv preprint arXiv:2405.21060}, 2024.

\bibitem{ozccelik2024chemical}
R. Özçelik, S. de Ruiter, E. Criscuolo, and F. Grisoni, ``Chemical language modeling with structured state space sequence models,'' \emph{Nat. Commun.}, vol. 15, no. 1, p. 6176, 2024.

\bibitem{shao2024deepseekmath}
Z. Shao, P. Wang, Q. Zhu, R. Xu, J. Song, X. Bi, \emph{et al.}, ``Deepseekmath: Pushing the limits of mathematical reasoning in open language models,'' \emph{arXiv preprint arXiv:2402.03300}, 2024.

\bibitem{guo2025deepseek}
D. Guo, D. Yang, H. Zhang, J. Song, R. Zhang, R. Xu, Q. Zhu, S. Ma, P. Wang, X. Bi, \emph{et al.}, ``Deepseek-r1: Incentivizing reasoning capability in LLMs via reinforcement learning,'' \emph{arXiv preprint arXiv:2501.12948}, 2025.

\bibitem{mroueh2025reinforcement}
Y. Mroueh, ``Reinforcement learning with verifiable rewards: GRPO's effective loss, dynamics, and success amplification,'' \emph{arXiv preprint arXiv:2503.06639}, 2025.

\bibitem{yan2022real}
S. Yan, Z. Wu, J. Wang, Y. Huang, M. Tan, and J. Yu, ``Real-world learning control for autonomous exploration of a biomimetic robotic shark,'' \emph{IEEE Trans. Ind. Electron.}, vol. 70, no. 4, pp. 3966--3974, 2022.

\bibitem{feng2025decentralized}
Y. Feng, Z. Wu, J. Wang, J. Gu, F. Yu, J. Yu, and M. Tan, ``Decentralized multirobotic fish pursuit control with attraction-enhanced reinforcement learning,'' \emph{IEEE Trans. Ind. Electron.}, vol. 72, no. 8, pp. 8290--8300, 2025.

\bibitem{feng2025cooperative}
Y.-K. Feng, Z.-X. Wu, and M. Tan, ``Cooperative pursuit policy for bionic underwater robot based on MARL-MHSA architecture: Data-driven modeling and distributed strategy optimization,'' \emph{Acta Autom. Sin.}, vol. 51, no. 9, pp. 1001--1014, 2025.




\end{thebibliography}
\end{document}